%% file: latex/sample-sigconf.tex
\documentclass[sigconf]{acmart}

\usepackage{graphicx}

\usepackage{booktabs}
\usepackage{multirow}
\usepackage{multicol}
\usepackage{amsmath}
\usepackage{amsfonts}

\AtBeginDocument{%
  }

\setcopyright{acmlicensed}

\copyrightyear{2026}
\acmYear{2026}
\setcopyright{cc}
\setcctype{by}
\acmConference[KDD 2026] {Proceedings of the 32nd ACM SIGKDD Conference on Knowledge Discovery and Data Mining V.2}{August 9--13, 2026}{Jeju Island, Republic of Korea.}
\acmBooktitle{Proceedings of the 32nd ACM SIGKDD Conference on Knowledge Discovery and Data Mining V.2 (KDD 2026), August 9--13, 2026, Jeju Island, Republic of Korea}
\acmISBN{979-8-4007-2259-2/2026/08}
\acmDOI{10.1145/3770855.3818113}

\begin{document}

\title{Training-free Truthfulness Detection via Sparse MLP Value Vectors}

\author{Runheng Liu}
\authornote{Both authors contributed equally to this research.}
\affiliation{%
  \institution{Beijing Institute of Technology}
  \city{Beijing}
  \country{China}}
\email{rhliu@bit.edu.cn}

\author{Heyan Huang}
\affiliation{%
  \institution{Beijing Institute of Technology}
  \city{Beijing}
  \country{China}}
\email{hhy63@bit.edu.cn}

\author{Xingchen Xiao}
\authornotemark[1]
\affiliation{%
  \institution{Beijing Institute of Technology}
  \city{Beijing}
  \country{China}}
\email{xcxiao@bit.edu.cn}

\author{Yanghao Zhou}
\affiliation{%
  \institution{Beijing Institute of Technology}
  \city{Beijing}
  \country{China}}
\email{zhouyh77@bit.edu.cn}

\author{Zhijing Wu}
\authornote{Corresponding author.}
\affiliation{%
  \institution{Beijing Institute of Technology}
  \city{Beijing}
  \country{China}}
\email{wuzhijing.joyce@gmail.com}

\renewcommand{\shortauthors}{Runheng Liu, Heyan Huang, Xingchen Xiao, Yanghao Zhou, and Zhijing Wu}

\input{latex/00_abstract}




\maketitle
\newcommand\kddavailabilityurl{https://doi.org/10.5281/zenodo.20390783}
\ifdefempty{\kddavailabilityurl}{}{
\begingroup\small\noindent\raggedright\textbf{Resource Availability:}\\
The source code of this paper has been made publicly available at \url{\kddavailabilityurl}.
\endgroup
}

\input{latex/01_introduction}
\input{latex/02_related_works}
\input{latex/03_preliminaries}
\input{latex/04_investigation}

\input{latex/05_truthv}

\input{latex/06_conclusion}

\begin{acks}
We thank all the anonymous reviewers for their insightful and valuable feedback on this paper. This work is supported by the National Natural Science Foundation of China (Grant No.U21B2009).
\end{acks}

\bibliographystyle{ACM-Reference-Format}
\bibliography{custom}

\input{latex/0A_appendix}

\end{document}

%% file: latex/00_abstract.tex
\begin{abstract}

Large language models (LLMs) are prone to generating factually incorrect content, motivating methods for assessing truthfulness from internal model signals. While supervised probing approaches can be effective, they require labeled data and classifier training. Recent training-free methods avoid parameter optimization but rely on coarse activation statistics that provide limited insight into how truthfulness-related signals arise within the model.
We present a training-free approach that operates at the level of individual multi-layer perceptron (MLP) value vectors. Through a systematic analysis, we find that although most value vectors show no meaningful signal, a sparse subset exhibits stable and directionally consistent correlations with content truthfulness. Leveraging this observation, we propose \textbf{TruthV}, a simple inference method that aggregates preferences expressed by these value vectors. TruthV requires only a small support set to identify relevant vectors and introduces no additional model parameters or classifier weights. We evaluate TruthV across model scales from 2B to 13B and multiple benchmarks, including question answering, natural language understanding, and hallucination evaluation. TruthV consistently outperforms existing training-free baselines, demonstrating that truthfulness-related variation in LLMs is captured in a sparse and structured manner at the level of MLP value vectors.

\end{abstract}

%% file: latex/01_introduction.tex
\input{figures/fig_show}
\section{Introduction}

Large Language Models (LLMs) have demonstrated strong empirical performance across a wide range of natural language tasks \cite{brown2020languagemodelsfewshotlearners,openai2024gpt4technicalreport}. Despite these advances, LLMs are known to generate factually incorrect or unsupported statements, commonly referred to as hallucinations. Such behavior poses challenges for reliability and safety, motivating substantial interest in methods for assessing the truthfulness of model-generated content.

Most existing approaches formulate truthfulness detection as a supervised probing problem, where classifiers are trained on LLM's internal activations to predict output truthfulness \cite{azaria-mitchell-2023-internal,marks2024the,li2023inferencetime,liu-etal-2024-universal,yuksekgonul2024attention,chuang-etal-2024-lookback}. While these studies demonstrate that internal representations can encode useful signals, probing can requires substantial labeled data and classifier training \cite{du2024haloscope,park2025steer}. Moreover, it often struggles to generalize to out-of-distribution (OOD) data samples and may require additional training when applied to new tasks \cite{burns2023discovering,marks2024the,ch-wang-etal-2024-androids,liu-etal-2024-universal}.

Independent of parameter training, recent work has proposed training-free approaches that identify truthfulness-related signals directly from pretrained model activations. In particular, NoVo \cite{ho2025novo} detects truthfulness by selecting a small set of attention heads and comparing the $\ell_2$ norms of their output vectors for the final token. While empirically effective and avoiding classifier training, the signal it exploits is inherently coarse: each attention norm aggregates the output of an entire head into a single scalar, providing a summary of overall activation magnitude without revealing how individual components respond to the content. Consequently, although such norms can correlate with truthfulness, they do not allow examination of specific directions or parameters, limiting the ability to analyze fine-grained, content-level correlations between activations and truthfulness.

To address this limitation, we focus on the multi-layer perceptron (MLP) modules in Transformer architectures. Each MLP layer contains a large set of value vectors that independently transform contextualized token representations. Prior work has shown that these vectors can form key–value memory structures, encoding factual and semantic information \cite{geva-etal-2021-transformer,geva-etal-2022-transformer,dai-etal-2022-knowledge}. Importantly, the activation of each vector defines an independent direction in representation space that can be examined in isolation, providing a structured and interpretable substrate for identifying correlations between model activations and content truthfulness.

Motivated by this architectural property, we conduct a systematic investigation of individual MLP value vectors across multiple multiple-choice question (MCQ) datasets. We analyze the activation of each value vector for the final token under different candidate answers, treating the content corresponding to the correct answer as truthful. Our findings reveal a sparse subset of vectors with consistent, directionally stable behavior: activation magnitudes reliably increase or decrease depending on whether the content is truthful. These patterns emerge naturally from the pretrained model and do not require classifier training or learned thresholds. We denote these recurring behaviors as \emph{argmax} and \emph{argmin} patterns.

Building on these observations, we propose TruthV, a training-free method for truthfulness detection. TruthV treats each selected value vector as an individual predictor, ranking candidate answers according to whether they maximize or minimize the vector's activation, and aggregates these rankings across vectors to produce a final prediction. By using only a small support set to select relevant vectors, TruthV avoids retraining model parameters or classifiers, making it flexible and scalable across tasks and model sizes.

We evaluate TruthV across a wide range of model scales (2B to 13B) and three benchmarks, including multiple-choice question answering, natural language understanding, and hallucination evaluation tasks. Across all settings, TruthV consistently outperforms existing training-free baselines while relying on the same small support set and no learned classifiers. These results demonstrate that truthfulness-related variation in pretrained language models is captured in a structured and robust manner at the level of individual MLP value vectors, enabling reliable content-level assessment without parameter training. In particular, their effectiveness on hallucination evaluation tasks demonstrates that the identified patterns generalize beyond simple question-answering scenarios, providing evidence that these activation-based signals reflect broader aspects of content reliability in LLM outputs.

%% file: figures/fig_show.tex
\begin{figure}[h]
    \centering
    \includegraphics[width=\linewidth]{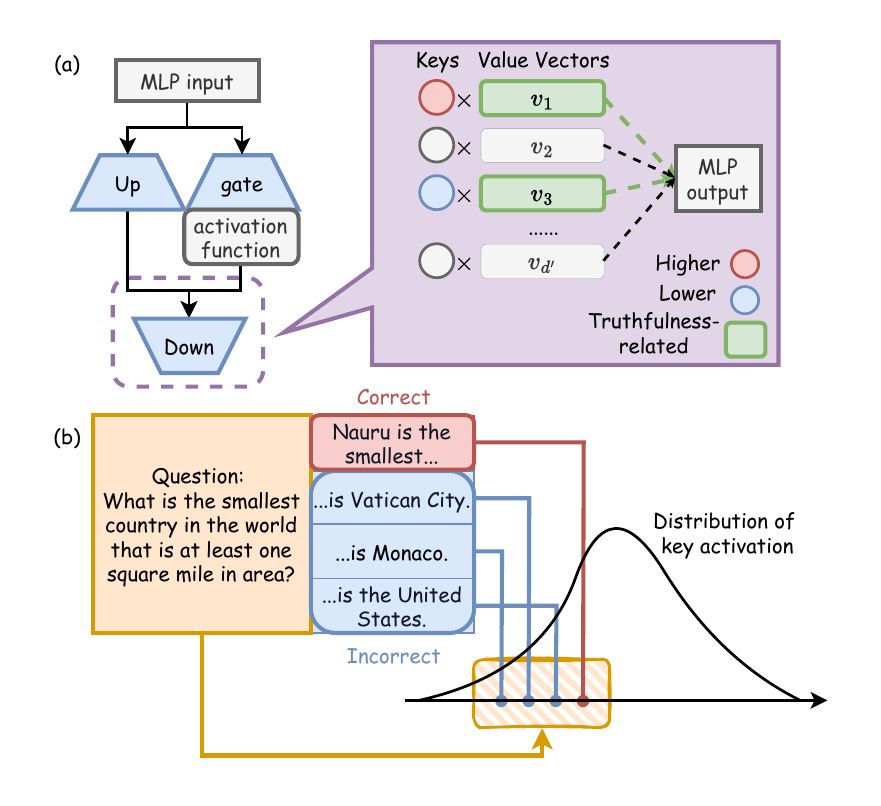}
    \caption{(a) illustrates the data flow through the MLP module when processing truthful content, which trigger certain key activations to be higher or lower. (b) shows an example of the argmax activation pattern where the correct answer results in a higher key activation compared to incorrect answers. The example is taken from the TruthfulQA (MC1) dataset. }
    \label{fig:show}
\end{figure}

%% file: latex/02_related_works.tex
\section{Related Works}

\paragraph{Truthfulness detection via internal activations.}

A growing body of work studies truthfulness in large language models by analyzing their internal activations. Many approaches formulate this problem as a supervised probing task, training classifiers on hidden representations to predict whether a model output is truthful \cite{alain2017understanding,azaria-mitchell-2023-internal,marks2024the,li2023inferencetime}. These probes have been applied to layer outputs, attention outputs, and attention weights, demonstrating that internal representations contain signals correlated with truthfulness \cite{li2023inferencetime,liu-etal-2024-universal,yuksekgonul2024attention,chuang-etal-2024-lookback}.  
More recently, training-free approaches have been proposed that avoid fitting explicit classifiers. NoVo \cite{ho2025novo} identifies a small set of attention heads whose output norms correlate with truthfulness and uses these statistics to perform inference with a limited support set. This line of work shows that truthfulness-related information can be extracted directly from pretrained model activations without optimizing additional parameters. Our work follows this training-free paradigm and investigates a different architectural locus for such signals.

\paragraph{Analysis of MLP modules in Transformers.}

The multi-layer perceptron (MLP) modules in Transformer architectures have been widely studied as meaningful computational components beyond their role as simple feedforward layers. \citet{geva-etal-2021-transformer} propose interpreting MLPs as key–value memory systems, where each column of the output projection matrix functions as a value vector activated by a learned key. Building on this view, \citet{dai-etal-2022-knowledge} show that key activations in BERT correlate with factual knowledge expression in cloze-style tasks. \citet{geva-etal-2022-transformer} further demonstrate that individual value vectors in GPT-2 encode interpretable concepts and influence specific token predictions.
These studies establish that MLP value vectors encode structured semantic and factual information and can be meaningfully analyzed at the level of individual parameters. While prior work primarily examines token-level associations and interpretability, our work investigates whether value-level activations also exhibit systematic correlations with content-level properties such as truthfulness, and whether these correlations can be exploited in a training-free manner.

%% file: latex/03_preliminaries.tex
\section{Preliminaries}

\paragraph{MLP module in LLMs.}

The MLP module is applied after the multi-head attention module in each Transformer layer, serving to project representations to a higher-dimensional space and to introduce nonlinearity through activation functions. Recent LLMs commonly employ Gated Linear Units (GLUs) and their variants as activation functions \cite{shazeer2020gluvariantsimprovetransformer}. 
For example, Llama-2 \cite{touvron2023llama2openfoundation}, Llama-3 \cite{grattafiori2024llama3herdmodels} and Qwen3 \cite{yang2025qwen3technicalreport} adopt SwiGLU, and Gemma-2 \cite{gemmateam2024gemma2improvingopen} employs GeGLU. With GLU-based activations, the output of the MLP module is given by
\begin{align}\label{eq:glu}
    m=&W_{\text{down}}(f(W_{\text{gate}}h)\odot (W_{\text{up}}h)),
\end{align}
where $h\in \mathbb{R}^{d}$ is the input to the MLP, $f$ denotes an element-wise activation function, and $\odot$ denotes the Hadamard product. Here, $W_{\text{gate}}, W_{\text{up}} \in \mathbb{R}^{d'\times d}$ and $W_{\text{down}} \in \mathbb{R}^{d \times d'}$, where typically $d' > d$.

\paragraph{Memory interpretation for MLP.}

\citet{geva-etal-2021-transformer} suggests that the MLP module can be interpreted as a key-value memory, where each column of the final projection matrix $W_{\text{down}}$ acts as a value vector, and the corresponding activation serves as a key. Under this view, Eq. \ref{eq:glu} can be reformulated as
\begin{equation}\label{eq:memory}
    m=\sum_{i=1}^{d'}k_iv_i,
\end{equation}
where $k_i=f(w_{\text{gate},i}^Th)(w_{\text{up},i}^Th)$ and $v_i=w_{\text{down},i}$. Here, $w_{\text{gate},i}, w_{\text{up},i} \in \mathbb{R}^{d}$ are the $i$th rows of $W_{\text{gate}}$ and $W_{\text{up}}$, respectively, and $w_{\text{down},i} \in \mathbb{R}^{d}$ is the $i$th column of $W_{\text{down}}$.

%% file: latex/04_investigation.tex
\input{figures/fig_ranking_of_v}
\section{Identifying Truthfulness-related Value Vectors}

To investigate whether certain value vectors within MLP modules are associated with content truthfulness, we focus on the task of multiple-choice question (MCQ), where a model must select the correct answer from a set of candidates. For each question $q$ with $M$ candidate answers $\{a_i\}_{i=1}^M$, we construct input sequences $\{[q; a_i]\}_{i=1}^M$, where the prefix $q$ is shared and the variation in truthfulness is confined to the answer suffix. These sequences are then fed into the model to analyze how key activations vary with respect to the truthfulness of the input.

\subsection{Settings}

\paragraph{Dataset.} 

We used data from NoVo benchmark \cite{ho2025novo}, which includes various MCQ datasets covering diverse topics, formats, and reasoning abilities. These datasets comprise TruthfulQA MC1 (TQA) \cite{lin-etal-2022-truthfulqa} for evaluating truthfulness, CommonsenseQA 2.0 (CSQA2) \cite{talmor2022commonsenseqa20exposinglimits} for commonsense reasoning, QASC \cite{Khot_Clark_Guerquin_Jansen_Sabharwal_2020} for scientific knowledge, SWAG \cite{zellers-etal-2018-swag} and HellaSwag (HSwag) \cite{zellers-etal-2019-hellaswag} for challenging commonsense sentence completions, SIQA \cite{sap-etal-2019-social} for social reasoning, PIQA \cite{Bisk_Zellers_Le_bras_Gao_Choi_2020} for physical reasoning, CosmosQA (Cosmos) \cite{huang-etal-2019-cosmos} for causal reasoning in narratives, and CICERO V1 and V2 (CICv1, CICv2) \cite{ghosal-etal-2022-cicero,shen2022multiviewcontextualcommonsenseinference} for multi-turn and strategic reasoning in dialogues. For each dataset, NoVo provides an optimal instruction along with 30 labeled samples for exploitation. These samples are drawn from the training set of each dataset, except for TQA, which lacks a training set; its samples are taken from the  ARC-Easy training set \cite{clark2018thinksolvedquestionanswering}, making TQA a zero-shot case. 
Following the NoVo' setting, we prepend the given instruction to $[q,a_i]$, ensuring a uniform content format across examples. We used the full training set of each dataset except for TQA during this experiment.

\paragraph{Models.}

We utilize various modern instruction-tuned LLMs in the experiment, including Llama-2-7B-Chat \cite{touvron2023llama2openfoundation}, Gemma-2-2B-it \cite{gemmateam2024gemma2improvingopen}, Llama-3.2-3B-Instruct \cite{grattafiori2024llama3herdmodels}, and Qwen3-4B \cite{yang2025qwen3technicalreport}.

\paragraph{Activation patterns triggered by truthful content.}

We hypothesize that certain value vectors in the MLP modules encode information relevant to truthfulness, and that their corresponding key activations are strongly modulated depending on whether the content is truthful. Considering the formulation of the scalar key activation $k_i=f(w_{\text{gate},i}^Th)(w_{\text{up},i}^Th)$ (see Eq. \ref{eq:memory}), such modulation can result in either the highest or the lowest activation among candidate answers for a given question. We refer to these two cases as exhibiting \textit{argmax} and \textit{argmin} activation patterns, respectively.

\paragraph{Scoring and ranking value vectors.}

Based on these activation patterns, we evaluate the ability of each value vector to distinguish correct answers in order to identify those most aligned with truthfulness. Following prior work \cite{li2023inferencetime,marks2024the,ho2025novo}, we focus on the final token of each input sequence. For the argmax pattern, we compute the accuracy of each value vector across the dataset $\mathcal{D}$ as:
\begin{equation}\label{eq:acc}
    \frac{1}{|\mathcal{D}|}\sum_{i=1}^{|\mathcal{D}|}\left[(\arg\max_j k_{i,j})=l_i\right],
\end{equation}
where $k_{i,j}$ denotes the key activation for the $j$th candidate in the $i$th MCQ, and $l_i$ is the index of the correct answer. We rank all value vectors according to this score and select the top $p$ as the most strongly aligned with the argmax activation pattern. A symmetric procedure is used to identify value vectors aligned with the argmin activation pattern by selecting those whose key activations are minimized by the correct answer.

\subsection{Results}

\paragraph{Existence of truthfulness-related value vectors.}

We rank MLP value vectors by their accuracy and consistently observe that a small subset achieves performance substantially above the random baseline. Figure~\ref{fig:ranking_of_v} shows results for several model--dataset pairs. While most value vectors perform close to chance, a minority attain notably higher accuracy under the argmax pattern. Each vector’s accuracy is computed by averaging over at least 1000 multiple-choice questions, making it highly unlikely for such performance to arise from random guessing alone.
To illustrate this quantitatively, consider the QASC dataset with eight answer choices. The probability that a random predictor achieves an accuracy of at least 0.351 on 1000 questions is
$\sum_{k=351}^{1000} \binom{1000}{k} \left(\tfrac{1}{8}\right)^k \left(\tfrac{7}{8}\right)^{1000-k} \approx 6.3 \times 10^{-76}$. This probability is vanishingly small, indicating that the high-performing value vectors observed in practice cannot be explained by random fluctuation.
In addition to high-performing vectors under the argmax pattern, we also observe a subset whose accuracy falls well below the random baseline. When evaluating these vectors under the argmin pattern using the same argmax-based ranking, their performance increases substantially (see the orange line in Figure~\ref{fig:ranking_of_v}). This consistent inversion suggests that these vectors exhibit stable directional preferences with respect to correct versus incorrect answer choices. Together, these results demonstrate that a small subset of MLP value vectors displays systematic, non-random activation behavior correlated with answer correctness.

\input{figures/fig_stat_layers_of_v}
\input{figures/fig_key_distribution_piqa}

\paragraph{Layer-wise distribution of truthfulness-related value vectors.}

We next examine where in the model these value vectors are concentrated. For each dataset, we select the top 0.1\% of value vectors by accuracy, assuming that these are the most indicative of truthfulness, based on the above analysis. We then aggregate these value vectors and analyze their distribution across the layers of each model. As shown in Figure \ref{fig:stat_layers_of_v}, these value vectors are predominantly located in the middle and final layers, and this trend holds consistently across different models. The distributions for both argmax and argmin patterns are highly similar, again highlighting their underlying complementarity. Notably, we observe no such vectors in the early layers, indicating that truthfulness-related modulation is largely absent from the early processing stages and begins to emerge in the middle layers. This observation aligns with previous work showing that internal activations from later layers are especially informative for truthfulness detection \cite{li2023inferencetime,marks2024the,azaria-mitchell-2023-internal,burns2023discovering,chuang-etal-2024-lookback}.

\paragraph{Semantic interpretability of truthfulness-related value vectors.}

Previous work has shown that some value vectors in transformer models encode human-interpretable concepts, such as the names of foods or drinks \cite{geva-etal-2022-transformer}. These vectors can be projected into the vocabulary space, enabling a rough approximation of their semantics by identifying the tokens with which they are most closely aligned. Based on this perspective, we select the value vector with the highest accuracy for each dataset and assess whether it exhibits similar interpretability. Following \citet{geva-etal-2022-transformer}, the value vector $v$ is projected into the vocabulary space via $r = Ev \in \mathbb{R}^{|\mathcal{V}|}$, where $E$ is the output embedding matrix of the model and $|\mathcal{V}|$ is the vocabulary size. We then analyze the top-10 tokens with the highest projection scores. In particular, we find no evidence of surface-level semantic alignment between the top-scoring tokens and the concept of truthfulness. This contrasts with the findings in GPT-2 under similar projection-based analysis, where many value vectors were aligned with interpretable concepts \cite{geva-etal-2022-transformer}. However, even in that study, approximately 50\% of value vectors could not be matched to any identifiable concept. It is therefore plausible that the selected truthfulness-related value vectors belong to this non-interpretable subset, which may encode more abstract or distributed properties. In addition, differences in model architecture—such as the use of GLU-based MLPs in modern LLMs compared to the GELU-based ones in GPT-2—could further reduce the direct interpretability of individual value vectors. Furthermore, t-SNE visualizations of these selected value vectors reveal no clear clustering, further suggesting that they do not form a semantically coherent subspace.

\paragraph{Context-dependent key activations of truthfulness-related value vectors.}

We further analyze the key activations associated with truthfulness-related value vectors to understand how their signals vary across different inputs. Although these vectors exhibit systematic differences in activation magnitude between truthful and untruthful answers under the argmax pattern, it is not evident that a single global threshold can reliably separate the two classes across all inputs. 
We conduct this analysis on the PIQA dataset, a two-choice question answering benchmark that enables controlled comparisons between truthful and untruthful content. Specifically, we select the value vector with the highest accuracy on the training set and examine its key activations on the same data. Figure~\ref{fig:key_distribution:piqa} shows the aggregated distributions of key activations for truthful and untruthful inputs. Despite the strong discriminative performance of this vector, the two distributions exhibit substantial overlap—a pattern that is consistently observed across different models. 
Importantly, this overlap arises when activations are aggregated across many different questions. At the level of individual questions, the key activation for the truthful answer is consistently higher (or lower, depending on the activation pattern) than that of the untruthful alternative. However, different questions induce different baseline activation ranges, resulting in a wide spread of values when pooled together. As a result, although the relative ordering between truthful and untruthful answers is preserved within each question, their absolute activation magnitudes vary across questions, leading to overlapping global distributions.
This analysis indicates that the key activations of truthfulness-related value vectors are context-dependent: the question prompt establishes a question-specific activation range, within which the candidate answers induce systematic but relative differences. Consequently, truthfulness is reflected in comparative activation patterns rather than in absolute activation thresholds that generalize across inputs.

%% file: figures/fig_ranking_of_v.tex
\begin{figure*}[tbp]
    \centering
    \includegraphics[width=\linewidth]{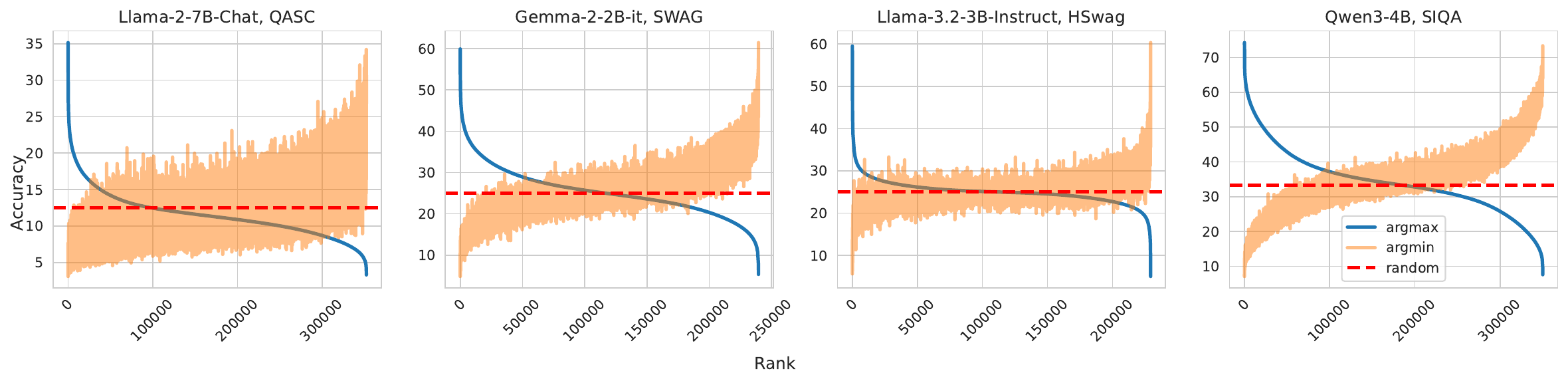}
    \caption{Accuracy of value vectors under both argmax and argmin activation patterns, ranked by argmax accuracy. The red dash line denotes the random guess result of each dataset.
    }
    \label{fig:ranking_of_v}
\end{figure*}

%% file: figures/fig_stat_layers_of_v.tex
\begin{figure*}[thb!]
    \centering
    \includegraphics[width=\linewidth]{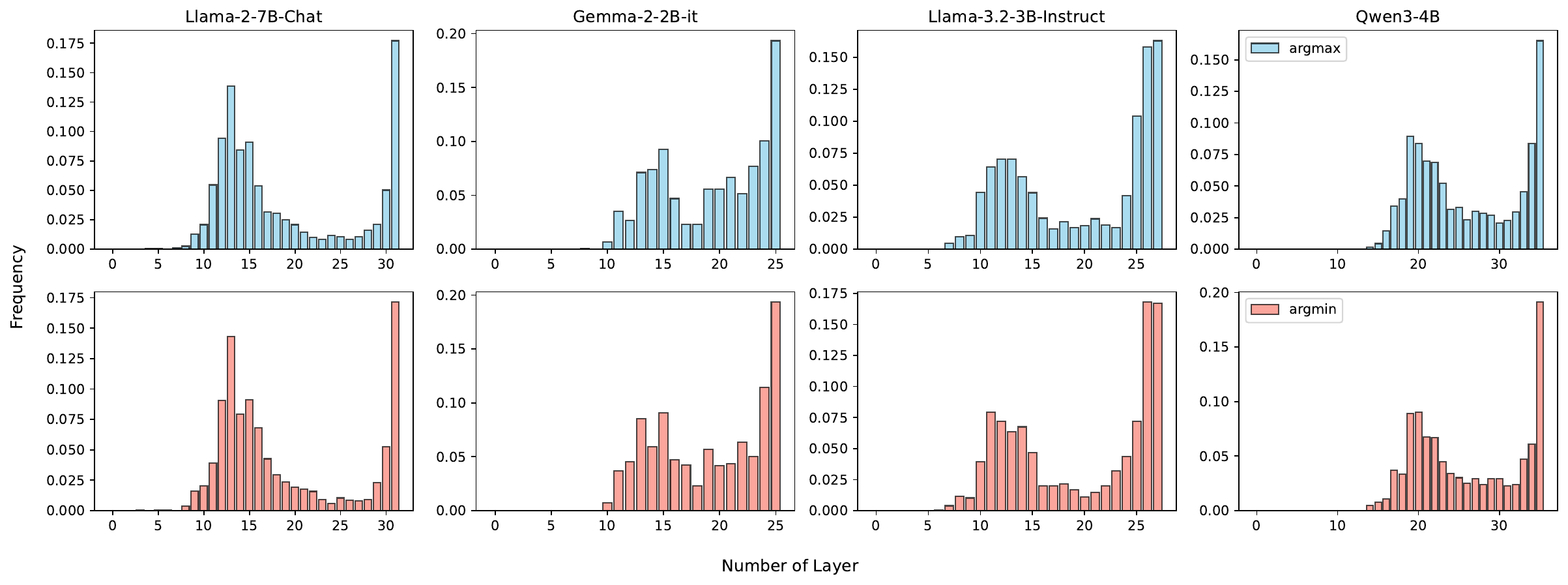}
    \caption{Layer-wise distribution of the top 0.01\% most accurate value vectors across different LLMs.}
    \label{fig:stat_layers_of_v}
\end{figure*}

%% file: figures/fig_key_distribution_piqa.tex
\begin{figure*}[thbp!]
    \centering
    \includegraphics[width=\linewidth]{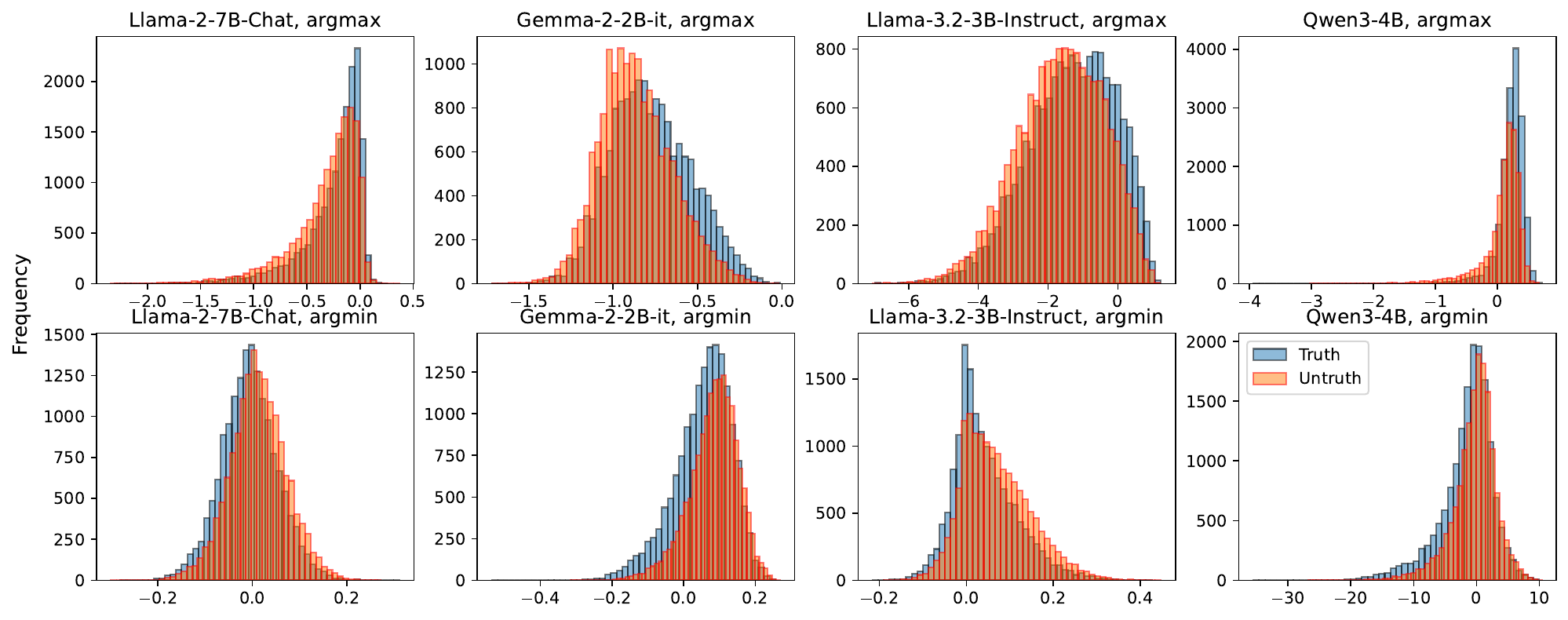}
    \caption{Key activation distributions of the most truthfulness-predictive value vector on the PIQA dataset.}
    \label{fig:key_distribution:piqa}
\end{figure*}

%% file: latex/05_truthv.tex
\section{TruthV: Training-free Truthfulness Detection via Value Vectors in LLM}

Our previous analysis has demonstrated the existence of truthfulness-related value vectors in LLMs. Building on these findings, we next examine how these value vectors can be employed to detect the truthfulness of content.

\subsection{Settings}

\paragraph{Datasets.}

We utilize data from NoVo benchmark \cite{ho2025novo}. In contrast to the previous section, which used the full training set to select truthfulness-related value vectors for analysis, we adopt the 30 provided examples, for fair comparison with baseline methods. More experiments on the Adversarial GLUE (AdvGLUE) \cite{wang2021adversarial} and HaluEval \cite{li-etal-2023-halueval} benchmarks are detailed in the Appendix \ref{app:addtional_exp}.

\paragraph{Method.}

Previous studies have proposed ensembling various truthfulness predictors to improve prediction accuracy \cite{ho2025novo,ch-wang-etal-2024-androids,himmi-etal-2024-enhanced}. In our settings, each value vector acts as an individual predictor. We select the top-$p$ value vectors based on their individual performance on the 30 provided samples and ensemble their predictions. Since these predictions are not expressed as normalized probability distributions, we follow the methodology of \citet{ho2025novo}, by computing the mode of the individual predictions. For the argmax pattern, the $c$-th value vector predicts an answer by choosing the candidate with the highest key activation: $a^c = \arg\max_j k^c_j$, where $k^c_j$ is the key activation corresponding to the $c$-th value vector when applied to the $j$th input $[q;a_j]$. We aggregate the predictions $\{a^c\}$ via majority voting:
\begin{equation}
    \hat{a} = \arg\max_{a \in M} \text{Count}(a, \{a^c\}),
\end{equation}
where $M$ is the set of all candidate answer indices, and $\text{Count}(a, \cdot)$ returns the number of times answer $a$ appears among the predictions. The final predicted answer $\hat{a}$ is the one that receives the most votes. A symmetric procedure is applied for the argmin pattern.

\paragraph{Baselines.}

We compare our method against two baselines: NoVo and Log-Likelihood. NoVo leverages the 30 provided samples to select a subset of attention heads. It uses the norm of their output to make predictions, and the predictions from the top-$p$ heads are aggregated via voting to obtain the final prediction. The optimal value of $p$ is empirically set to 85\%. The Log-Likelihood baseline simply chooses the answer with the highest log-likelihood score.

\paragraph{Metrics.}

Following the setting of the benchmark, we report accuracy as the evaluation metrics.

\paragraph{Implementation details.}

All experiments were conducted on an NVIDIA A100-40G GPU.

\input{tables/tab_novo_bench}

\subsection{Main Results}

In Table~\ref{tab:novo_bench}, we present the overall performance of TruthV and the baseline methods on the NoVo benchmark. TruthV consistently achieves the best average accuracy across all four LLMs, outperforming existing baselines by a significant margin. For instance, under the argmax pattern, it achieves an average accuracy of 70.33 on Gemma-2-2B-it, representing an improvement of 8.66 points over the previous best of 61.67. These gains are consistently observed across both the argmax and argmin patterns, which show only slight differences in accuracy. Moreover, TruthV outperforms baselines on 9 out of 10 datasets, which span diverse topics, formats, and reasoning abilities, highlighting its robustness across tasks.

\input{figures/fig_p}

\subsection{Ablations}

\paragraph{Effect of the number of selected value vectors.}

We examine the impact of the hyperparameter $p$, which controls the number of value vectors included in the ensemble by varying it from 0.01\% to 1\%. As shown in Figure \ref{fig:p}, performance rises rapidly with increasing $p$, peaks at a relatively small value, and then gradually declines, with only a modest drop afterward. We hypothesize that a very small $p$ leads to overfitting on the 30 labeled samples used for selection, while an excessively large $p$ introduce unrelated value vectors that dilute useful signals. Both activation patterns exhibit similar trend across different $p$ values, suggesting consistent behavior with respect to ensemble size. In addition, we observe that the optimal value of $p$ varies across LLMs. To identify a unified setting, we compute the average accuracy across all four models and find that the performance peaks around $p=0.001$. Accordingly, we adopt $p=0.001$ for all experiments.

\paragraph{Effect of activation pattern.}

Table~\ref{tab:novo_bench} has shown that the value vectors selected under both argmax and argmin activation patterns can detect the truthfulness of the content. Here, we examine whether combining their predictions can lead to further improvements. Specifically, we aggregated the predictions from both activation patterns and performed majority voting over the combined sets. As shown in Figure \ref{fig:p}, the optimal ensemble size for the combined pattern follows a trend similar to that observed for the individual argmax and argmin patterns. Moreover, the combined pattern slightly outperforms both individual patterns. This marginal improvement suggests that the value vectors selected by the two activation patterns largely encode overlapping information but also contribute minor complementary signals.

\input{tables/tab_novo_scale_samples}

\subsection{Analysis}

\paragraph{Efficiency of selection.}

The results of the above experiments demonstrate that even with only 30 samples, the selected value vectors yield competitive performance. Intuitively, incorporating more samples into the selection process should improve the performance of TruthV. Therefore, we further use the full training set of each dataset for selection (except for TQA, which lacks a training set). Since training set sizes vary across datasets, we search $p$ from 0.01\% to 1\% and report the best result for each dataset. We adopt the argmax activation pattern throughout this experiment. As shown in Table \ref{tab:novo_scale_samples}, using the full training set consistently leads to better performance compared to using only 30 samples. However, in a few settings (e.g., Llama-2-7B-Chat on SWAG), the improvement is marginal or even slightly negative. This suggests that many truthfulness-related value vectors are already ranked near the top with just 30 samples, so adding more data provides limited benefit. In most scenarios, while using more data does improve the ranking of truthfulness-related value vectors, TruthV already achieves competitive results even under the low-resource setting, suggesting that these value vectors are salient and can be reliably identified with limited supervision.

\subsection{Cross-Dataset Transfer of Truthfulness-Related Value Vectors}

\input{tables/tab_novo_generalization}

We examine how value vectors identified by TruthV perform when transferred across datasets. For each source dataset, we select relevant value vectors using the full training set under the argmax activation pattern, and evaluate them on the validation sets of other target datasets without any additional training or adaptation. As shown in Table~\ref{tab:novo:generalization}, the selected vectors achieve performance consistently above the random baseline across all target datasets. While cross-dataset accuracy is generally lower than that obtained by selecting vectors using only 30 labeled samples from the target dataset, minimal target-specific supervision improves performance without requiring model retraining.
Consistent with prior work showing limited cross-dataset transfer in probing-based classifiers \cite{ch-wang-etal-2024-androids,liu-etal-2024-universal} and training-free methods such as NoVo \cite{ho2025novo}, TruthV exhibits a similar trend: transferring vectors across datasets leads to decreased accuracy, which can be partially mitigated with a small support set from the target dataset.

\subsection{Scalability of TruthV to Larger Models}\label{app:results_with_larger_models}

\input{tables/tab_novo_scale_model_size}

We examine whether the value-vector-based signals exploited by TruthV persist and remain effective as model capacity increases.

\paragraph{Models.}

We further extend our evaluation to larger models to assess the scalability of TruthV. Specifically, we consider Llama-2-13B-Chat \cite{touvron2023llama2openfoundation}, Llama-3.1-8B-Instruct \cite{grattafiori2024llama3herdmodels}, and Gemma-2-9B-it \cite{gemmateam2024gemma2improvingopen}. 
Consistent with our main experiments, we report results using the argmax activation pattern, which remains dominant across model scales.

\paragraph{Results.}

As shown in Table~\ref{tab:novo_scale_model_size}, TruthV consistently outperforms both NoVo and the log-likelihood baseline across all evaluated datasets and model scales. This indicates that the truth-correlated value vectors identified by TruthV are not specific to small models, but persist and remain effective as model capacity increases. Moreover, within the same model family, larger models consistently yield stronger performance. For example, TruthV achieves an accuracy of 78.36 on Gemma-2-9B-it compared to 70.33 on Gemma-2-2B-it, and 70.23 on Llama-2-13B-Chat compared to 65.98 on Llama-2-7B-Chat. This trend suggests that as models scale, truthfulness-related signals in MLP value vectors become more pronounced and easier to exploit in a training-free manner. Importantly, this improvement does not rely on additional supervision or learned parameters. Instead, it reflects a strengthening of structured activation patterns within the MLP modules themselves, supporting our central claim that MLP value vectors provide a scalable and reliable substrate for content-level truthfulness assessment.

\subsection{Generalization to Natural Language Understanding Tasks}

\input{tables/tab_adv_glue_bench}

\paragraph{Datasets.}

We evaluate the generalization of TruthV on the Adversarial GLUE (AdvGLUE) benchmark \cite{wang2021adversarial}, which is designed to test model robustness on challenging natural language understanding tasks. Following the training-free setting, we randomly select 30 samples from the validation split as a support set to identify truth-correlated value vectors, and use the remaining samples for evaluation. We repeat this process with random seeds from 42 to 51, resulting in 10 independent runs, and report the mean accuracy. We adopt the task-specific instructions used in the NoVo benchmark. Since NoVo does not specify a concrete input--output format for AdvGLUE, we design a unified format and apply it consistently to all methods for fair comparison. For example, in the MNLI subtask, we format each input as "premise: \{premise\} hypothesis: \{hypothesis\}", and use the label names "entailment", "neutral", and "contradiction" as candidate answers. All training-free baselines are evaluated using the same support sets, prompts, and evaluation protocol as TruthV.

\paragraph{Results.}

Table~\ref{tab:adv_glue_bench} summarizes the performance of TruthV and baseline methods on AdvGLUE. Across all four LLMs, TruthV achieves the highest average accuracy, consistently outperforming existing training-free baselines. For instance, under the argmax activation pattern, TruthV reaches an average accuracy of 72.59 on Qwen3-4B, improving upon the strongest baseline by 14.62 points on average. Notably, the improvements hold for both argmax and argmin activation patterns, and TruthV outperforms baselines on every dataset in the benchmark. These results indicate that the truth-correlated activation patterns identified in MLP value vectors generalize beyond multiple-choice question answering, extending to diverse natural language understanding tasks under adversarial settings.

\subsection{Generalization to Hallucination Evaluation Task}

\input{tables/tab_halu_bench}

\paragraph{Datasets.}

We further evaluate TruthV on the HaluEval benchmark \cite{li-etal-2023-halueval}, a large-scale benchmark designed to assess hallucinations in model-generated responses. HaluEval covers multiple generation settings, including question answering, knowledge-grounded dialogue, and text summarization, and focuses on determining whether a generated response is factually supported by the input context.
To apply TruthV in this setting, we reformulate hallucination detection as a binary decision problem over candidate labels \emph{Yes} (non-hallucinated) and \emph{No} (hallucinated). Specifically, we treat the task prompt and the model-generated response as the input, and evaluate the activation patterns of MLP value vectors for the final token under each candidate label. For example, in the summarization task, the input is formatted as ``Document: \{document\} Summarization: \{summary\}'', where correct summaries are labeled as \emph{Yes} and hallucinated summaries as \emph{No}. This formulation allows us to directly test whether truth-correlated activation patterns identified in MLP value vectors extend to settings where the input includes free-form generated text.
Following the training-free protocol, we randomly select 30 samples as a support set to identify truth-correlated value vectors and randomly sample 1,000 additional instances for evaluation. We repeat this procedure with random seeds from 42 to 51, resulting in 10 independent runs, and report mean accuracy. All training-free baselines are evaluated using the same prompts, support sets, and evaluation protocol.

\paragraph{Results.}

Table~\ref{tab:halueval_bench} reports the performance of TruthV and baseline methods on HaluEval. Across all four LLMs, TruthV consistently achieves the highest average accuracy, substantially outperforming existing training-free baselines. For example, under the argmax activation pattern, TruthV attains an average accuracy of 81.07 on Qwen3-4B, exceeding the strongest baseline by 10.04 points on average.
Importantly, these improvements hold across all HaluEval subsets, including question answering, dialogue, and summarization, and for both argmax and argmin activation patterns. These results demonstrate that the activation patterns identified in MLP value vectors are not limited to discriminative or classification-style tasks, but also generalize to detecting hallucinations in model-generated content. This provides evidence that TruthV captures content-level reliability signals that remain effective in realistic generation settings.

%% file: tables/tab_novo_bench.tex
\begin{table*}[t]
    \centering
    \caption{The overall performance of various methods on the NoVo benchmark. For each dataset, the best results are marked in \textbf{bold}.}
    \setlength{\tabcolsep}{1mm}
    \resizebox{\textwidth}{!}{
    \begin{tabular}{cccccccccccc|c}
    \toprule
         Model&  Method     &  TQA&CQA2& QASC&  SWAG& HSwag&SIQA& PIQA& Cosmos&CICv1 &CICv2 &Avg.\\
         \midrule
 \multirow{4}{*}{Llama-2-7B-Chat}& Log-Likelihood& 34.88& \textbf{55.73}& 17.49& 48.67& 47.57& 46.37& 75.08& 25.43& 31.43&34.28 &41.69\\

 & NoVo  &  70.01&55.41& 43.63& 68.67& 59.80& 60.03& 69.75& 52.19& 35.91&62.79 &57.82\\

 & TruthV(argmax)& 70.13& 53.68& \textbf{57.02}& \textbf{75.43}& 69.67 & \textbf{66.38}& \textbf{77.04}& 62.35& \textbf{53.07}&\textbf{74.98}&\textbf{65.98}\\
 & TruthV(argmin)& \textbf{71.97}& 53.88& 54.43& 74.92& \textbf{70.85}& 65.25& 75.68& \textbf{66.23}& 50.34& 73.49&65.70\\
 
 \midrule
 \multirow{4}{*}{Gemma-2-2B-it}& Log-Likelihood& 47.00& \textbf{58.84}& 60.26& 53.28& 48.21& 49.13& 76.66& 29.48& 32.90&36.78 &49.25\\
 & NoVo  &  63.16&56.20& 51.73& 59.63& 59.22& 67.76& 73.94& 65.19& 50.44&69.39 &61.67\\
 & TruthV(argmax)& 69.40& 56.55& \textbf{73.97}& \textbf{73.40}& 69.85& \textbf{69.86}& 79.92& \textbf{71.66}& \textbf{60.78}&77.87&\textbf{70.33}\\
 & TruthV(argmin)& \textbf{71.48}& 55.37& 73.54& 71.15& \textbf{70.79}& 69.81& \textbf{80.14}& 71.42& 60.72& \textbf{78.05}&70.25\\
 \midrule
 \multirow{4}{*}{Llama-3.2-3B-Instruct}& Log-Likelihood& 39.29& 58.76& \textbf{53.35}& 49.84& 48.30& 49.28& \textbf{74.37}& 30.12& 29.12&33.82 &46.62\\
 & NoVo  &  58.02&\textbf{59.78}& 40.71& \textbf{61.75}& 61.18& 55.17& 66.65& 44.62& 27.80&47.04 &52.27\\
 & TruthV(argmax)& \textbf{60.71}& 59.74&  50.11& 59.35& \textbf{68.15}& \textbf{61.31}& 72.63& \textbf{53.23}& \textbf{37.20}&\textbf{57.84}&\textbf{58.03}\\
 & TruthV(argmin)& 55.45& 59.62& 39.63& 58.70& 66.24& 60.70& 69.48& 52.86& 37.01& 55.10&55.48\\
 \midrule
 \multirow{4}{*}{Qwen3-4B}& Log-Likelihood&  35.74&\textbf{64.94}& 44.92& 44.47& 36.13& 39.76& 67.79& 22.61& 21.08&32.36 &40.98\\
 & NoVo  &  \textbf{68.67}&57.50& 66.74& 68.30& 59.69& 65.61& 73.07& 59.60& 43.14&70.31 &63.26\\
 & TruthV(argmax)& 67.32& 60.57& \textbf{76.24}& \textbf{74.42}& \textbf{71.34}& \textbf{69.04}& \textbf{80.90}& \textbf{65.96}& 59.40&\textbf{77.62}&\textbf{70.28}\\
 & TruthV(argmin)& 67.81& 60.72& 75.70& 73.96& 70.28& 68.78& 80.36& 63.92& \textbf{59.44}& 76.87&69.78\\
 \bottomrule
    \end{tabular}
    }
    \label{tab:novo_bench}
\end{table*}

%% file: figures/fig_p.tex
\begin{figure*}[thbp!]
    \centering
    \includegraphics[width=\linewidth]{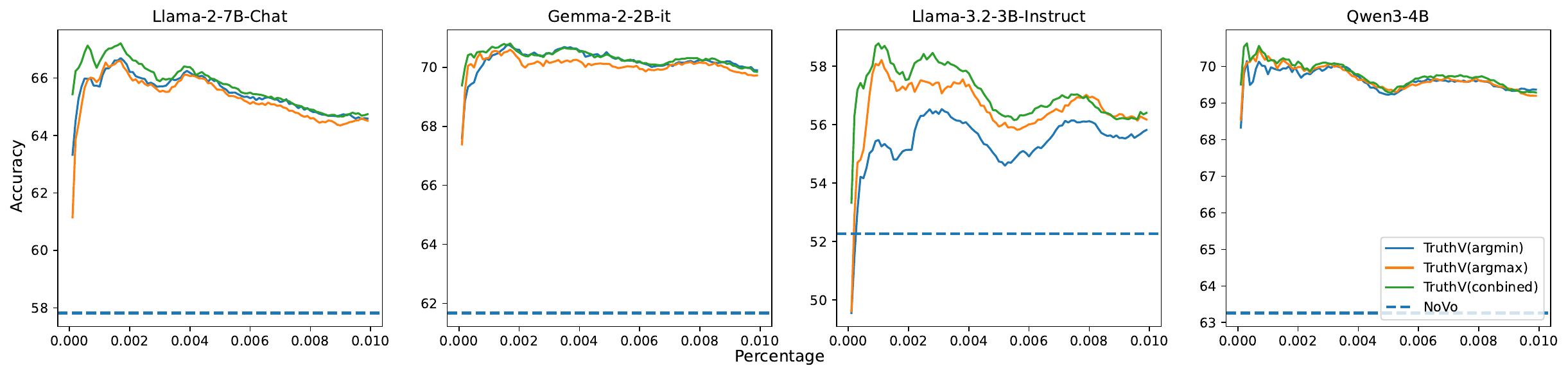}
    \caption{Performance of TruthV across varying $p$, where the accuracy is averaged over all datasets.}
    \label{fig:p}
\end{figure*}

%% file: tables/tab_novo_scale_samples.tex
\begin{table*}[t]
    \centering
    \caption{Performance comparison of different budget in selecting truthfulness-related value vectors, under the argmax activation pattern. }
    \begin{tabular}{ccccccccccc}
    \toprule
         Model &Budget &CQA2& QASC&  SWAG& HSwag&SIQA& PIQA& Cosmos&CICv1&CICv2  \\
         \midrule
    
 \multirow{2}{*}{Llama-2-7B-Chat}&30&53.68& 57.02& \bf 75.43& 69.67& 66.38& 77.04& 62.35& 53.07& 74.98  \\
  &All& \bf 58.40& \bf 64.79& 75.32& \bf 70.69& \bf 68.78& \bf 79.82& \bf 69.51& \bf 60.43& \bf 79.15\\
  \midrule
  \multirow{2}{*}{Gemma-2-2B-it}&30&56.55& 73.97& 73.40& 69.85& 69.86& 79.92& 71.66& 60.78& 77.87  \\
  &All& \bf 59.35& \bf 74.73& \bf 74.61& \bf 73.84& \bf 71.14& \bf 81.01& \bf 72.50& \bf 65.14& \bf 79.54\\
  \midrule
  \multirow{2}{*}{Llama-3.2-3B-Instruct}&30&59.74& 50.11& 59.35& 68.15& 61.31& 72.63& 53.23& 37.20& 57.84  \\
  &All& \bf 61.71& \bf 71.60& \bf 70.82& \bf 71.52& \bf 67.04& \bf 75.57& \bf 65.86& \bf 53.23& \bf 76.16\\
  \midrule
  \multirow{2}{*}{Qwen3-4B}&30&60.57& 76.24& 74.42& 71.34& 69.04& \bf 80.90& 65.96& 59.40& 77.62  \\
  &All& \bf 63.60& \bf 78.40& \bf 75.45& \bf 72.42& \bf 72.11& 80.58& \bf 67.71& \bf 63.83&   \bf 80.29\\
 \bottomrule
    \end{tabular}
    \label{tab:novo_scale_samples}
\end{table*}

%% file: tables/tab_novo_generalization.tex
\begin{table}[h!]
    \centering
    \caption{Cross-dataset evaluation of value vectors selected on source datasets and evaluated on target datasets, using Gemma-2-2B-it under the argmax activation pattern.}
    \setlength{\tabcolsep}{1mm}
    \resizebox{0.5\textwidth}{!}{
    \begin{tabular}{cccccccccc}
    \toprule
         Source$\downarrow$Target$\rightarrow$&CQA2& QASC&  SWAG& HSwag&SIQA& PIQA& Cosmos&CICv1 &CICv2  \\
    \midrule
 CQA2&\textbf{59.35}& 63.07& 65.67& 63.88& 67.81& 75.46& 64.29& 42.82& 57.77 \\
 QASC&58.52& \textbf{74.73}& 69.39& 67.19& 68.63& 80.25& 68.51& 48.83& 60.01 \\
 SWAG&58.60& 71.92& \textbf{74.61}& 72.29& 68.63& 80.69& 67.07& 52.24& 64.93 \\
 HSwag&58.91& 71.49& 72.31&  \textbf{73.84}& 68.78& 80.58& 65.16& 53.79& 63.08 \\
 SIQA&58.13& 64.36& 67.94& 67.49& \textbf{71.14}& 79.11& 64.89& 53.20& 71.13 \\
 PIQA&58.36& 69.11& 70.33& 70.91& 69.91& 81.01& 63.99& 54.88& 68.96 \\
 Cosmos&57.30& 73.43& 72.01& 70.87& 70.98& \textbf{81.45}& \textbf{72.50}& 57.23& 69.53 \\
 CICv1 &58.13& 68.36& 73.83& 71.06& 70.68& 80.30& 62.58& \textbf{65.14}& 73.95 \\
 CICv2&56.55& 55.29& 64.94& 63.56& 70.27& 80.09& 61.57& 61.72& \textbf{79.54}\\
 \midrule
 Random Guess&50.00& 12.50& 25.00& 25.00& 33.33& 50.00& 25.00& 20.00& 25.00\\
 \bottomrule
    \end{tabular}
    }
    \label{tab:novo:generalization}
\end{table}

%% file: tables/tab_novo_scale_model_size.tex
\begin{table*}[t]
    \centering
    \caption{Scalability of TruthV across larger backbone models on the NoVo benchmark. For each dataset, the best results are marked in \textbf{bold}.}
    \setlength{\tabcolsep}{1mm}
    \resizebox{\textwidth}{!}{
    \begin{tabular}{cccccccccccc|c}
    \toprule
         Model&  Method     &  TQA&CQA2& QASC&  SWAG& HSwag&SIQA& PIQA& Cosmos&CICv1 &CICv2 &Avg.\\
         \midrule
 \multirow{3}{*}{Llama-3.1-8B-Instruct}& Log-Likelihood& 42.96& 65.21& 55.18& 53.84& 55.23& 53.48& \bf 79.43& 28.68& 32.91&36.24&50.32\\

 & NoVo  &  58.63&66.51& 54.54& 69.06& 66.35& 64.64& 69.80& 53.70& 40.86&59.66&60.38\\

 & TruthV(argmax)& \bf 68.05& \bf 68.87& \bf 70.41& \bf 73.27& \bf 78.81& \bf 70.93& 79.11& \bf 72.83& \bf 51.69&\bf 66.79&\bf 70.08\\
 
 \midrule
 \multirow{3}{*}{Gemma-2-9B-it}& Log-Likelihood& 43.82& \bf 72.02& 66.63& 54.77& 51.73& 51.28& 77.48& 30.39& 34.29&38.10&52.05\\
 & NoVo  &  78.09&69.78& 78.83& 75.35& 70.98& 72.47& 81.77& 75.24& 47.73&69.03&71.93\\
 & TruthV(argmax)& \bf 79.68& 71.74& \bf 84.45& \bf 79.65& \bf 80.97& \bf 75.84& \bf 87.21& \bf 77.82& \bf 65.42&\bf 80.83&\bf 78.36\\
 \midrule
 \multirow{3}{*}{Llama-2-13B-Chat}& Log-Likelihood& 38.68& 56.87& 26.67& 49.78& 51.85& 49.18& 75.79& 23.02& 34.89&35.03&44.18\\

 & NoVo  &  \bf 76.62&57.50& 55.08& 66.59& 68.12& 65.10& 72.52& 54.51& 38.94&72.84&62.78\\

 & TruthV(argmax)& 70.38& \bf 58.13& \bf 65.12& \bf 73.45& \bf 78.35& \bf 70.78& \bf 79.60& \bf 69.38& \bf 57.00&\bf 80.15&\bf 70.23\\
 \bottomrule
    \end{tabular}
    }
    \label{tab:novo_scale_model_size}
\end{table*}

%% file: tables/tab_adv_glue_bench.tex
\begin{table}[th]
    \centering
    \caption{The overall performance of various methods on the NoVo benchmark. For each dataset, the best results are marked in \textbf{bold}.}
    \resizebox{0.5\textwidth}{!}{
    \begin{tabular}{cccccccc|c}
    \toprule
         Model&  Method      & SST2&QQP &MNLI & MNLI-MM& QNLI& RTE &Avg.\\
         \midrule
 \multirow{4}{*}{Llama-2-7B-Chat}& Log-Likelihood & 42.57& 41.03& 42.98& 36.42& 50.00&  43.21&42.70\\

 & NoVo   & 61.95& 56.67& 39.78& 32.95& 52.54&  57.06&50.16\\

 & TruthV(argmax)& \bf 71.61& 79.38& 52.42& 41.06& 64.75&  \bf 74.90&\bf 64.02\\
 & TruthV(argmin)& 70.93& \bf 79.58& \bf 52.86& \bf 41.21& \bf 65.25&  73.33&63.86\\
 
 \midrule
 \multirow{4}{*}{Gemma-2-2B-it}& Log-Likelihood & 52.70& 41.03& 61.16& 43.83& 50.00& 43.21 &48.65\\
 & NoVo   & 65.25& 59.38& 54.07& 38.33& 53.05& 61.18 &55.21\\
 & TruthV(argmax) & 71.10& 82.50& 74.62& 52.65& 69.92& 80.59 &71.90\\
 & TruthV(argmin) & \bf 71.27& \bf 82.71& \bf 74.84& \bf 53.11& \bf 70.08&  \bf 81.57&\bf 72.26\\
 \midrule
 \multirow{4}{*}{Llama-3.2-3B-Instruct}& Log-Likelihood & 63.51& 41.03& 32.23& 27.78& 50.00& 43.21 &42.96\\
 & NoVo   & 63.47& 58.75& 40.33& 34.39& 50.00& 55.88 &50.47\\
 & TruthV(argmax) & \bf 65.34& \bf 84.58& \bf 50.22& \bf 51.89& \bf 62.88& 74.90 &64.97\\
 & TruthV(argmin) & 64.92& 82.29& 50.11& 51.82& \bf 62.88& \bf 78.82&\bf 65.14\\
 \midrule
 \multirow{4}{*}{Qwen3-4B}& Log-Likelihood & 53.38& 41.03& 45.45& 37.04& 50.00&  43.21&45.02\\
 & NoVo   & 61.53& 58.54& 52.31& 41.36& 65.08&  69.02&57.97\\
 & TruthV(argmax) & \bf 65.85& \bf 79.17& \bf 78.24& \bf 52.80& \bf 77.54&  \bf 81.96&\bf 72.59\\
 & TruthV(argmin) & 65.08& 77.71& \bf 78.24& \bf 52.80& 77.29&  81.76&72.15\\
 \bottomrule
    \end{tabular}
    }
    \label{tab:adv_glue_bench}
\end{table}

%% file: tables/tab_halu_bench.tex
\begin{table}[t]
    \centering
    \caption{Performance of various methods on the HaluEval benchmark. For each dataset, the best results are marked in \textbf{bold}.}
    \setlength{\tabcolsep}{1mm}
    \resizebox{0.5\textwidth}{!}{
    \begin{tabular}{ccccc|c}
    \toprule
         Model&  Method     &   QA&Dialogue&Summarization&Avg.\\
         \midrule
 \multirow{4}{*}{Llama-2-7B-Chat}& Log-Likelihood& 48.93& 49.38& 52.49&50.27\\
 & NoVo  & 65.51& 56.94& 56.27&59.57\\
 & TruthV(argmax)& 79.71& 57.28& \bf 61.37&\bf 66.12\\
 & TruthV(argmin)& \bf 80.20& \bf 57.35& 60.61&66.05\\
 \midrule
        
 \multirow{4}{*}{Gemma-2-2B-it}& Log-Likelihood&  67.34&59.23& 52.43&59.67\\
 & NoVo  &   78.88&62.06&60.52&67.15\\

 & TruthV(argmax)&  \bf 89.63&68.36& 63.87&73.95\\
 & TruthV(argmin)& 89.57& \bf 68.89& \bf 64.23&\bf 74.23\\
 \midrule
 \multirow{4}{*}{Llama-3.2-3B-Instruct}& Log-Likelihood& 53.27& 54.32& 51.70&53.10\\
 & NoVo  & 70.99& 61.67& 57.08&63.25\\
 & TruthV(argmax)& \bf 93.89& \bf 67.59& 59.10&\bf 73.53\\
  & TruthV(argmin)& 93.74& 67.40& \bf 59.18&73.44\\
  \midrule
 \multirow{4}{*}{Qwen3-4B}& Log-Likelihood& 46.99& 59.42& 45.52&50.64\\
 & NoVo  & 78.42& 67.22& 67.45&71.03\\
 & TruthV(argmax)& \bf 89.58& \bf 70.02& \bf 83.60&\bf 81.07\\
  & TruthV(argmin)& 89.28& 69.43& 83.50&80.74\\
 \bottomrule
    \end{tabular}
    }
    \label{tab:halueval_bench}
\end{table}

%% file: latex/06_conclusion.tex
\section{Conclusion}

We investigated how truthfulness-related signals are reflected in the internal activations of pretrained language models and introduced \textbf{TruthV}, a training-free approach that leverages individual MLP value vectors for truthfulness detection. Our study shows that a sparse subset of value vectors exhibits systematic and directionally consistent activation differences between truthful and untruthful content, enabling reliable prediction without training classifiers or modifying model parameters.
Across multiple benchmarks and model scales, TruthV achieves strong performance compared to existing training-free baselines while relying on the same small support set. Beyond empirical gains, our analysis provides evidence that truthfulness-related variation in LLM outputs is captured in a structured manner at the level of individual MLP value vectors, offering a more fine-grained perspective on internal representations than previously explored activation statistics. 
We hope this work encourages further investigation of value-level representations as a useful lens for understanding reliability-related behaviors in large language models, and for developing lightweight methods that operate directly on pretrained model internals.

%% file: latex/0A_appendix.tex
\appendix

\section{Details of Benchmarks}

\input{tables/tab_details_novo}
\input{tables/tab_details_advglue}
\input{tables/tab_details_halueval}

\section{Extended Evaluation and Analysis}\label{app:addtional_exp}

\subsection{Are TruthV and NoVo Complementary?}

\input{tables/tab_ensemble}
\input{figures/fig_ensemble_on_qasc}
\input{tables/tab_novo_mask}
Since TruthV and NoVo are both training-free methods that rely on internal activations but operate on different architectural components, a natural question is whether their signals are complementary. To examine this question in a simple and diagnostic manner, we perform a score-level ensembling experiment that combines the outputs of TruthV and NoVo.

Specifically, for each method, we first compute a scalar score for each candidate answer. We then normalize these scores across all candidates by dividing each score by the sum of scores over the candidate set, i.e.,
\[
\tilde{s}_i = \frac{s_i}{\sum_j s_j},
\]
which maps the scores to the range $[0,1]$ while preserving their relative magnitudes. The normalized scores from TruthV and NoVo are then summed, and the candidate with the highest combined score is selected. We conduct this experiment using Gemma-2-2B-it as the backbone model and evaluate both the argmax and argmin activation patterns.

As shown in Table~\ref{tab:ensemble}, ensembling TruthV with NoVo consistently results in a performance decline compared to using TruthV alone, across both activation patterns. This outcome suggests that the signals exploited by TruthV and NoVo are not directly complementary under naive score-level combination, and that simple averaging introduces interference rather than additive gains.

We further investigate the effect of the ensemble ratio. As shown in Figure~\ref{fig:ensemble_on_qasc}, varying the weight assigned to TruthV and NoVo does not yield performance improvements over TruthV alone. In particular, increasing the contribution of NoVo consistently degrades accuracy. This trend indicates that the attention-based norm signals used by NoVo may conflict with the value-level activation patterns identified by TruthV, rather than reinforcing them.

These observations are consistent with our earlier analysis that TruthV appears to capture a more structured and fine-grained signal at the level of individual MLP value vectors, whereas NoVo relies on aggregated attention-based statistics. When combined without explicit coordination, these heterogeneous signals can dilute the discriminative patterns leveraged by TruthV. We leave the exploration of more principled fusion strategies that jointly model attention- and MLP-based signals to future work.

\subsection{Specificity Analysis via Content Neutralization}

To examine whether the identified value vectors respond to truth-related semantic content rather than generic surface or syntactic patterns, we conduct a content-neutralization experiment. Specifically, we construct a masked version of the MCQ datasets by replacing all nouns and proper nouns with a unified placeholder token \texttt{[NOUN]}. This transformation removes most factual content while largely preserving sentence structure, length, and grammatical form.
We evaluate the value vectors selected under the argmax activation pattern on Gemma-2-2B-it using these masked inputs. As shown in Table~\ref{tab:novo_mask}, performance drops substantially compared to the original unmasked setting. This degradation indicates that the discriminative signal exploited by these value vectors is not preserved when factual content is removed, and therefore cannot be attributed solely to surface-level or syntactic regularities.
Notably, accuracy on masked inputs remains slightly above the random baseline. This residual performance is expected, as multiple-choice datasets often contain structural biases (e.g., answer length or phrasing patterns) that persist even after content masking. Nevertheless, the pronounced performance decline demonstrates that the activation patterns identified by TruthV rely primarily on content-dependent information, rather than generic linguistic cues.

%% file: tables/tab_details_novo.tex
\begin{table}[h]
    \centering
    \caption{Statistics of the NoVo Benchmark.}
    \begin{tabular}{cccc}
    \toprule
         Dataset&  Train & Test &Validation\\
         \midrule
 TruthfulQA MC1& - & -&684 \\
 ARC-Easy& 2251& 2376&570\\
 CommonsenseQA 2.0&  9264&  -&2541\\

 QASC&  8134&  920&926\\
 
 SWAG&  73546&  20006&20005\\
 HellaSwag&  39905&  10042&10003\\
 SIQA&  33410&  -&1954\\
 PIQA&  16113&  3084&1838\\
 CosmosQA&  25262&  6963&2985\\
 CICERO V1&  27225&  -&9470\\
 CICERO V2&  13496&  -&2806\\

 \bottomrule
    \end{tabular}
    \label{tab:details_novo}
\end{table}

%% file: tables/tab_details_advglue.tex
\begin{table}[h]
    \centering
    \caption{Statistics of the Adversarial GLUE Benchmark.}
    \begin{tabular}{cc}
    \toprule
         Dataset&Validation\\
         \midrule
 MNLI&121\\
 MNLI-MM&162\\
 QNLI&148\\

 QQP&78\\
 
 RTE&81\\
 SST2&148\\

 \bottomrule
    \end{tabular}
    \label{tab:details_novo}
\end{table}

%% file: tables/tab_details_halueval.tex
\begin{table}[h]
    \centering
    \caption{Statistics of the HaluEval Benchmark.}
    \begin{tabular}{cc}
    \toprule
         Dataset&Validation\\
         \midrule
  QA&10000\\
 Dialogue&10000\\
 Summarization&10000\\
 \bottomrule
    \end{tabular}
    \label{tab:details_halueval}
\end{table}

%% file: tables/tab_ensemble.tex
\begin{table*}[t]
    \centering
    \caption{Ensemble result of TruthV and NoVo, with Gemma-2-2B-it as the backbone.}
    \setlength{\tabcolsep}{1mm}
    \begin{tabular}{lcccccccccc|c}
    \toprule
           Method     &  TQA&CQA2& QASC&  SWAG& HSwag&SIQA& PIQA& Cosmos&CICv1 &CICv2 &Avg.\\
\midrule
    
  NoVo  &  63.16&56.20& 51.73& 59.63& 59.22& 67.76& 73.94& 65.19& 50.44&69.39 &61.67\\

  TruthV(argmax)& 69.40& 56.55& 73.97& 73.40& 69.85& 69.86& 79.92& 71.66& 60.78&77.87&70.33\\
  TruthV(argmin)& 71.48& 55.37& 73.54& 71.15& 70.79& 69.81& 80.14& 71.42& 60.72& 78.05&70.25\\
  \midrule
  TruthV(argmax)+NoVo& 68.18& 57.30& 69.44& 70.72& 67.20& 68.88& 77.80& 71.22& 57.60& 77.05&68.54\\
  TruthV(argmin)+NoVo& 70.01& 56.83& 70.84& 69.68& 67.67& 68.99& 78.35& 70.69& 57.18& 77.90&68.81\\
  \bottomrule
    \end{tabular}
    \label{tab:ensemble}
\end{table*}

%% file: figures/fig_ensemble_on_qasc.tex
\begin{figure}[thbp!]
    \centering
    \includegraphics[width=\linewidth]{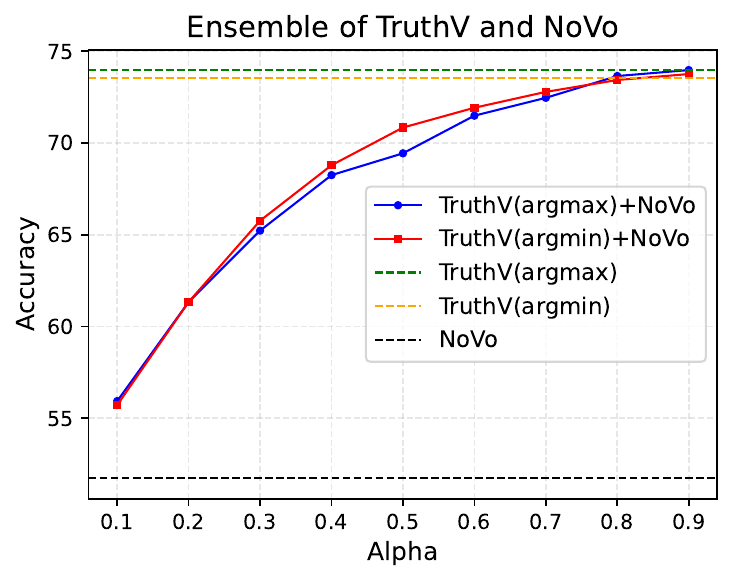}
    \caption{Performance of ensembling TruthV and NoVo on the QASC dataset under varying ensemble ratios $\alpha$, where the final score is computed as $\alpha \cdot s_{\text{TruthV}} + (1-\alpha) \cdot s_{\text{NoVo}}$ after score normalization. Gemma-2-2B-it is used as the backbone model.}

    \label{fig:ensemble_on_qasc}
\end{figure}

%% file: tables/tab_novo_mask.tex
\begin{table*}[thbp!]
    \centering
     \caption{Performance of TruthV on content-neutralized MCQs, where all nouns and proper nouns are replaced by a unified placeholder token, with Gemma-2-2B-it as the backbone.}
    \setlength{\tabcolsep}{1mm}
    \begin{tabular}{ccccccccccc|c}
    \toprule
           Method     &  TQA&CQA2& QASC&  SWAG& HSwag&SIQA& PIQA& Cosmos&CICv1 &CICv2 &Avg.\\
         \midrule
  TruthV(argmax)& 69.40& 56.55& 73.97& 73.40& 69.85& 69.86& 79.92& 71.66& 60.78&77.87&70.33\\
  TruthV(argmax)+Mask& 55.08& 48.92& 16.85& 38.38& 38.39& 44.47& 57.83& 44.59& 27.50&63.65&43.57\\
  Random& 25.00& 50.00& 12.50& 25.00& 25.00& 33.33& 50.00& 25.00& 20.00& 25.00&29.08\\
  \bottomrule
    \end{tabular}
    \label{tab:novo_mask}
\end{table*}

%% file: custom.bib
@inproceedings{
park2025steer,
title={Steer {LLM} Latents for Hallucination Detection},
author={Seongheon Park and Xuefeng Du and Min-Hsuan Yeh and Haobo Wang and Yixuan Li},
booktitle={Forty-second International Conference on Machine Learning},
year={2025},
url={https://openreview.net/forum?id=UMqNQEPNT3}
}

@inproceedings{li-etal-2023-halueval,
    title = "{H}alu{E}val: A Large-Scale Hallucination Evaluation Benchmark for Large Language Models",
    author = "Li, Junyi  and
      Cheng, Xiaoxue  and
      Zhao, Xin  and
      Nie, Jian-Yun  and
      Wen, Ji-Rong",
    editor = "Bouamor, Houda  and
      Pino, Juan  and
      Bali, Kalika",
    booktitle = "Proceedings of the 2023 Conference on Empirical Methods in Natural Language Processing",
    month = dec,
    year = "2023",
    address = "Singapore",
    publisher = "Association for Computational Linguistics",
    url = "https://aclanthology.org/2023.emnlp-main.397/",
    doi = "10.18653/v1/2023.emnlp-main.397",
    pages = "6449--6464"
}

@inproceedings{
wang2021adversarial,
title={Adversarial {GLUE}: A Multi-Task Benchmark for Robustness Evaluation of Language Models},
author={Boxin Wang and Chejian Xu and Shuohang Wang and Zhe Gan and Yu Cheng and Jianfeng Gao and Ahmed Hassan Awadallah and Bo Li},
booktitle={Thirty-fifth Conference on Neural Information Processing Systems Datasets and Benchmarks Track (Round 2)},
year={2021},
url={https://openreview.net/forum?id=GF9cSKI3A_q}
}

@inproceedings{himmi-etal-2024-enhanced,
    title = "Enhanced Hallucination Detection in Neural Machine Translation through Simple Detector Aggregation",
    author = "Himmi, Anas  and
      Staerman, Guillaume  and
      Picot, Marine  and
      Colombo, Pierre  and
      Guerreiro, Nuno M",
    editor = "Al-Onaizan, Yaser  and
      Bansal, Mohit  and
      Chen, Yun-Nung",
    booktitle = "Proceedings of the 2024 Conference on Empirical Methods in Natural Language Processing",
    month = nov,
    year = "2024",
    address = "Miami, Florida, USA",
    publisher = "Association for Computational Linguistics",
    url = "https://aclanthology.org/2024.emnlp-main.1033/",
    doi = "10.18653/v1/2024.emnlp-main.1033",
    pages = "18573--18583"
}

@inproceedings{geva-etal-2022-transformer,
    title = "Transformer Feed-Forward Layers Build Predictions by Promoting Concepts in the Vocabulary Space",
    author = "Geva, Mor  and
      Caciularu, Avi  and
      Wang, Kevin  and
      Goldberg, Yoav",
    editor = "Goldberg, Yoav  and
      Kozareva, Zornitsa  and
      Zhang, Yue",
    booktitle = "Proceedings of the 2022 Conference on Empirical Methods in Natural Language Processing",
    month = dec,
    year = "2022",
    address = "Abu Dhabi, United Arab Emirates",
    publisher = "Association for Computational Linguistics",
    url = "https://aclanthology.org/2022.emnlp-main.3/",
    doi = "10.18653/v1/2022.emnlp-main.3",
    pages = "30--45"
}

@inproceedings{dai-etal-2022-knowledge,
    title = "Knowledge Neurons in Pretrained Transformers",
    author = "Dai, Damai  and
      Dong, Li  and
      Hao, Yaru  and
      Sui, Zhifang  and
      Chang, Baobao  and
      Wei, Furu",
    editor = "Muresan, Smaranda  and
      Nakov, Preslav  and
      Villavicencio, Aline",
    booktitle = "Proceedings of the 60th Annual Meeting of the Association for Computational Linguistics (Volume 1: Long Papers)",
    month = may,
    year = "2022",
    address = "Dublin, Ireland",
    publisher = "Association for Computational Linguistics",
    url = "https://aclanthology.org/2022.acl-long.581/",
    doi = "10.18653/v1/2022.acl-long.581",
    pages = "8493--8502"
}

@misc{
alain2017understanding,
title={Understanding intermediate layers using linear classifier probes},
author={Guillaume Alain and Yoshua Bengio},
year={2017},
url={https://openreview.net/forum?id=ryF7rTqgl}
}

@misc{openai2024gpt4technicalreport,
      title={GPT-4 Technical Report}, 
      author={OpenAI and Josh Achiam and Steven Adler and Sandhini Agarwal and Lama Ahmad and Ilge Akkaya and Florencia Leoni Aleman and Diogo Almeida and Janko Altenschmidt and Sam Altman and others},
      year={2024},
      eprint={2303.08774},
      archivePrefix={arXiv},
      primaryClass={cs.CL},
      url={https://arxiv.org/abs/2303.08774}, 
}

@misc{brown2020languagemodelsfewshotlearners,
      title={Language Models are Few-Shot Learners}, 
      author={Tom B. Brown and Benjamin Mann and Nick Ryder and Melanie Subbiah and Jared Kaplan and Prafulla Dhariwal and Arvind Neelakantan and Pranav Shyam and Girish Sastry and Amanda Askell and Sandhini Agarwal and Ariel Herbert-Voss and Gretchen Krueger and Tom Henighan and Rewon Child and Aditya Ramesh and Daniel M. Ziegler and Jeffrey Wu and Clemens Winter and Christopher Hesse and Mark Chen and Eric Sigler and Mateusz Litwin and Scott Gray and Benjamin Chess and Jack Clark and Christopher Berner and Sam McCandlish and Alec Radford and Ilya Sutskever and Dario Amodei},
      year={2020},
      eprint={2005.14165},
      archivePrefix={arXiv},
      primaryClass={cs.CL},
      url={https://arxiv.org/abs/2005.14165}, 
}

@inproceedings{geva-etal-2021-transformer,
    title = "Transformer Feed-Forward Layers Are Key-Value Memories",
    author = "Geva, Mor  and
      Schuster, Roei  and
      Berant, Jonathan  and
      Levy, Omer",
    editor = "Moens, Marie-Francine  and
      Huang, Xuanjing  and
      Specia, Lucia  and
      Yih, Scott Wen-tau",
    booktitle = "Proceedings of the 2021 Conference on Empirical Methods in Natural Language Processing",
    month = nov,
    year = "2021",
    address = "Online and Punta Cana, Dominican Republic",
    publisher = "Association for Computational Linguistics",
    url = "https://aclanthology.org/2021.emnlp-main.446/",
    doi = "10.18653/v1/2021.emnlp-main.446",
    pages = "5484--5495"
}

@misc{shazeer2020gluvariantsimprovetransformer,
      title={GLU Variants Improve Transformer}, 
      author={Noam Shazeer},
      year={2020},
      eprint={2002.05202},
      archivePrefix={arXiv},
      primaryClass={cs.LG},
      url={https://arxiv.org/abs/2002.05202}, 
}

@misc{yang2025qwen3technicalreport,
      title={Qwen3 Technical Report}, 
      author={An Yang and Anfeng Li and Baosong Yang and Beichen Zhang and Binyuan Hui and Bo Zheng and Bowen Yu and Chang Gao and Chengen Huang and Chenxu Lv and others},
      year={2025},
      eprint={2505.09388},
      archivePrefix={arXiv},
      primaryClass={cs.CL},
      url={https://arxiv.org/abs/2505.09388}, 
}

@misc{grattafiori2024llama3herdmodels,
      title={The Llama 3 Herd of Models}, 
      author={Aaron Grattafiori and Abhimanyu Dubey and Abhinav Jauhri and Abhinav Pandey and Abhishek Kadian and Ahmad Al-Dahle and Aiesha Letman and Akhil Mathur and Alan Schelten and Alex Vaughan and others},
      year={2024},
      eprint={2407.21783},
      archivePrefix={arXiv},
      primaryClass={cs.AI},
      url={https://arxiv.org/abs/2407.21783}, 
}

@misc{gemmateam2024gemma2improvingopen,
      title={Gemma 2: Improving Open Language Models at a Practical Size}, 
      author={Gemma Team and Morgane Riviere and Shreya Pathak and Pier Giuseppe Sessa and Cassidy Hardin and Surya Bhupatiraju and Léonard Hussenot and Thomas Mesnard and Bobak Shahriari and Alexandre Ramé and others},
      year={2024},
      eprint={2408.00118},
      archivePrefix={arXiv},
      primaryClass={cs.CL},
      url={https://arxiv.org/abs/2408.00118}, 
}

@misc{touvron2023llama2openfoundation,
      title={Llama 2: Open Foundation and Fine-Tuned Chat Models}, 
      author={Hugo Touvron and Louis Martin and Kevin Stone and Peter Albert and Amjad Almahairi and Yasmine Babaei and Nikolay Bashlykov and Soumya Batra and Prajjwal Bhargava and Shruti Bhosale and others},
      year={2023},
      eprint={2307.09288},
      archivePrefix={arXiv},
      primaryClass={cs.CL},
      url={https://arxiv.org/abs/2307.09288}, 
}

@misc{clark2018thinksolvedquestionanswering,
      title={Think you have Solved Question Answering? Try ARC, the AI2 Reasoning Challenge}, 
      author={Peter Clark and Isaac Cowhey and Oren Etzioni and Tushar Khot and Ashish Sabharwal and Carissa Schoenick and Oyvind Tafjord},
      year={2018},
      eprint={1803.05457},
      archivePrefix={arXiv},
      primaryClass={cs.AI},
      url={https://arxiv.org/abs/1803.05457}, 
}

@misc{shen2022multiviewcontextualcommonsenseinference,
      title={Multiview Contextual Commonsense Inference: A New Dataset and Task}, 
      author={Siqi Shen and Deepanway Ghosal and Navonil Majumder and Henry Lim and Rada Mihalcea and Soujanya Poria},
      year={2022},
      eprint={2210.02890},
      archivePrefix={arXiv},
      primaryClass={cs.CL},
      url={https://arxiv.org/abs/2210.02890}, 
}

@inproceedings{ghosal-etal-2022-cicero,
    title = "{CICERO}: A Dataset for Contextualized Commonsense Inference in Dialogues",
    author = "Ghosal, Deepanway  and
      Shen, Siqi  and
      Majumder, Navonil  and
      Mihalcea, Rada  and
      Poria, Soujanya",
    editor = "Muresan, Smaranda  and
      Nakov, Preslav  and
      Villavicencio, Aline",
    booktitle = "Proceedings of the 60th Annual Meeting of the Association for Computational Linguistics (Volume 1: Long Papers)",
    month = may,
    year = "2022",
    address = "Dublin, Ireland",
    publisher = "Association for Computational Linguistics",
    url = "https://aclanthology.org/2022.acl-long.344/",
    doi = "10.18653/v1/2022.acl-long.344",
    pages = "5010--5028"
}

@inproceedings{huang-etal-2019-cosmos,
    title = "Cosmos {QA}: Machine Reading Comprehension with Contextual Commonsense Reasoning",
    author = "Huang, Lifu  and
      Le Bras, Ronan  and
      Bhagavatula, Chandra  and
      Choi, Yejin",
    editor = "Inui, Kentaro  and
      Jiang, Jing  and
      Ng, Vincent  and
      Wan, Xiaojun",
    booktitle = "Proceedings of the 2019 Conference on Empirical Methods in Natural Language Processing and the 9th International Joint Conference on Natural Language Processing (EMNLP-IJCNLP)",
    month = nov,
    year = "2019",
    address = "Hong Kong, China",
    publisher = "Association for Computational Linguistics",
    url = "https://aclanthology.org/D19-1243/",
    doi = "10.18653/v1/D19-1243",
    pages = "2391--2401"
}

@article{Bisk_Zellers_Le_bras_Gao_Choi_2020, title={PIQA: Reasoning about Physical Commonsense in Natural Language}, volume={34}, url={https://ojs.aaai.org/index.php/AAAI/article/view/6239}, DOI={10.1609/aaai.v34i05.6239}, abstractNote={&lt;p&gt;To apply eyeshadow without a brush, should I use a &lt;em&gt;cotton swab or a toothpick&lt;/em&gt;? Questions requiring this kind of &lt;strong&gt;physical commonsense&lt;/strong&gt; pose a challenge to today’s natural language understanding systems. While recent pretrained models (such as BERT) have made progress on question answering over more &lt;em&gt;abstract&lt;/em&gt; domains – such as news articles and encyclopedia entries, where text is plentiful – in more &lt;em&gt;physical&lt;/em&gt; domains, text is inherently limited due to reporting bias. Can AI systems learn to reliably answer physical commonsense questions without experiencing the physical world?&lt;/p&gt;&lt;p&gt;In this paper, we introduce the task of physical commonsense reasoning and a corresponding benchmark dataset &lt;strong&gt;Physical Interaction: Question Answering&lt;/strong&gt; or &lt;strong&gt;PIQA&lt;/strong&gt;. Though humans find the dataset easy (95% accuracy), large pretrained models struggle (∼75%). We provide analysis about the dimensions of knowledge that existing models lack, which offers significant opportunities for future research.&lt;/p&gt;}, number={05}, journal={Proceedings of the AAAI Conference on Artificial Intelligence}, author={Bisk, Yonatan and Zellers, Rowan and Le bras, Ronan and Gao, Jianfeng and Choi, Yejin}, year={2020}, month={Apr.}, pages={7432-7439} }

@inproceedings{sap-etal-2019-social,
    title = "Social {IQ}a: Commonsense Reasoning about Social Interactions",
    author = "Sap, Maarten  and
      Rashkin, Hannah  and
      Chen, Derek  and
      Le Bras, Ronan  and
      Choi, Yejin",
    editor = "Inui, Kentaro  and
      Jiang, Jing  and
      Ng, Vincent  and
      Wan, Xiaojun",
    booktitle = "Proceedings of the 2019 Conference on Empirical Methods in Natural Language Processing and the 9th International Joint Conference on Natural Language Processing (EMNLP-IJCNLP)",
    month = nov,
    year = "2019",
    address = "Hong Kong, China",
    publisher = "Association for Computational Linguistics",
    url = "https://aclanthology.org/D19-1454/",
    doi = "10.18653/v1/D19-1454",
    pages = "4463--4473"
}

@inproceedings{zellers-etal-2019-hellaswag,
    title = "{H}ella{S}wag: Can a Machine Really Finish Your Sentence?",
    author = "Zellers, Rowan  and
      Holtzman, Ari  and
      Bisk, Yonatan  and
      Farhadi, Ali  and
      Choi, Yejin",
    editor = "Korhonen, Anna  and
      Traum, David  and
      M{\`a}rquez, Llu{\'i}s",
    booktitle = "Proceedings of the 57th Annual Meeting of the Association for Computational Linguistics",
    month = jul,
    year = "2019",
    address = "Florence, Italy",
    publisher = "Association for Computational Linguistics",
    url = "https://aclanthology.org/P19-1472/",
    doi = "10.18653/v1/P19-1472",
    pages = "4791--4800"
}

@inproceedings{zellers-etal-2018-swag,
    title = "{SWAG}: A Large-Scale Adversarial Dataset for Grounded Commonsense Inference",
    author = "Zellers, Rowan  and
      Bisk, Yonatan  and
      Schwartz, Roy  and
      Choi, Yejin",
    editor = "Riloff, Ellen  and
      Chiang, David  and
      Hockenmaier, Julia  and
      Tsujii, Jun{'}ichi",
    booktitle = "Proceedings of the 2018 Conference on Empirical Methods in Natural Language Processing",
    month = oct # "-" # nov,
    year = "2018",
    address = "Brussels, Belgium",
    publisher = "Association for Computational Linguistics",
    url = "https://aclanthology.org/D18-1009/",
    doi = "10.18653/v1/D18-1009",
    pages = "93--104"
}

@article{Khot_Clark_Guerquin_Jansen_Sabharwal_2020, title={QASC: A Dataset for Question Answering via Sentence Composition}, volume={34}, url={https://ojs.aaai.org/index.php/AAAI/article/view/6319}, DOI={10.1609/aaai.v34i05.6319}, abstractNote={&lt;p&gt;Composing knowledge from multiple pieces of texts is a key challenge in multi-hop question answering. We present a multi-hop reasoning dataset, &lt;strong&gt;Q&lt;/strong&gt;uestion &lt;strong&gt;A&lt;/strong&gt;nswering via &lt;strong&gt;S&lt;/strong&gt;entence &lt;strong&gt;C&lt;/strong&gt;omposition (QASC), that requires retrieving facts from a large corpus and composing them to answer a multiple-choice question. QASC is the first dataset to offer two desirable properties: (a) the facts to be composed are annotated in a large corpus, and (b) the decomposition into these facts is not evident from the question itself. The latter makes retrieval challenging as the system must introduce new concepts or relations in order to discover potential decompositions. Further, the reasoning model must then learn to identify valid compositions of these retrieved facts using common-sense reasoning. To help address these challenges, we provide annotation for supporting facts as well as their composition. Guided by these annotations, we present a two-step approach to mitigate the retrieval challenges. We use other multiple-choice datasets as additional training data to strengthen the reasoning model. Our proposed approach improves over current state-of-the-art language models by 11% (absolute). The reasoning and retrieval problems, however, remain unsolved as this model still lags by 20% behind human performance.&lt;/p&gt;}, number={05}, journal={Proceedings of the AAAI Conference on Artificial Intelligence}, author={Khot, Tushar and Clark, Peter and Guerquin, Michal and Jansen, Peter and Sabharwal, Ashish}, year={2020}, month={Apr.}, pages={8082-8090} }

@misc{talmor2022commonsenseqa20exposinglimits,
      title={CommonsenseQA 2.0: Exposing the Limits of AI through Gamification}, 
      author={Alon Talmor and Ori Yoran and Ronan Le Bras and Chandra Bhagavatula and Yoav Goldberg and Yejin Choi and Jonathan Berant},
      year={2022},
      eprint={2201.05320},
      archivePrefix={arXiv},
      primaryClass={cs.CL},
      url={https://arxiv.org/abs/2201.05320}, 
}

@inproceedings{lin-etal-2022-truthfulqa,
    title = "{T}ruthful{QA}: Measuring How Models Mimic Human Falsehoods",
    author = "Lin, Stephanie  and
      Hilton, Jacob  and
      Evans, Owain",
    editor = "Muresan, Smaranda  and
      Nakov, Preslav  and
      Villavicencio, Aline",
    booktitle = "Proceedings of the 60th Annual Meeting of the Association for Computational Linguistics (Volume 1: Long Papers)",
    month = may,
    year = "2022",
    address = "Dublin, Ireland",
    publisher = "Association for Computational Linguistics",
    url = "https://aclanthology.org/2022.acl-long.229/",
    doi = "10.18653/v1/2022.acl-long.229",
    pages = "3214--3252"
}

@inproceedings{ch-wang-etal-2024-androids,
    title = "Do Androids Know They`re Only Dreaming of Electric Sheep?",
    author = "CH-Wang, Sky  and
      Van Durme, Benjamin  and
      Eisner, Jason  and
      Kedzie, Chris",
    editor = "Ku, Lun-Wei  and
      Martins, Andre  and
      Srikumar, Vivek",
    booktitle = "Findings of the Association for Computational Linguistics: ACL 2024",
    month = aug,
    year = "2024",
    address = "Bangkok, Thailand",
    publisher = "Association for Computational Linguistics",
    url = "https://aclanthology.org/2024.findings-acl.260/",
    doi = "10.18653/v1/2024.findings-acl.260",
    pages = "4401--4420"
}

@inproceedings{chuang-etal-2024-lookback,
    title = "Lookback Lens: Detecting and Mitigating Contextual Hallucinations in Large Language Models Using Only Attention Maps",
    author = "Chuang, Yung-Sung  and
      Qiu, Linlu  and
      Hsieh, Cheng-Yu  and
      Krishna, Ranjay  and
      Kim, Yoon  and
      Glass, James R.",
    editor = "Al-Onaizan, Yaser  and
      Bansal, Mohit  and
      Chen, Yun-Nung",
    booktitle = "Proceedings of the 2024 Conference on Empirical Methods in Natural Language Processing",
    month = nov,
    year = "2024",
    address = "Miami, Florida, USA",
    publisher = "Association for Computational Linguistics",
    url = "https://aclanthology.org/2024.emnlp-main.84/",
    doi = "10.18653/v1/2024.emnlp-main.84",
    pages = "1419--1436"
}

@inproceedings{liu-etal-2024-universal,
    title = "On the Universal Truthfulness Hyperplane Inside {LLM}s",
    author = "Liu, Junteng  and
      Chen, Shiqi  and
      Cheng, Yu  and
      He, Junxian",
    editor = "Al-Onaizan, Yaser  and
      Bansal, Mohit  and
      Chen, Yun-Nung",
    booktitle = "Proceedings of the 2024 Conference on Empirical Methods in Natural Language Processing",
    month = nov,
    year = "2024",
    address = "Miami, Florida, USA",
    publisher = "Association for Computational Linguistics",
    url = "https://aclanthology.org/2024.emnlp-main.1012/",
    doi = "10.18653/v1/2024.emnlp-main.1012",
    pages = "18199--18224"
}

@inproceedings{
burns2023discovering,
title={Discovering Latent Knowledge in Language Models Without Supervision},
author={Collin Burns and Haotian Ye and Dan Klein and Jacob Steinhardt},
booktitle={The Eleventh International Conference on Learning Representations },
year={2023},
url={https://openreview.net/forum?id=ETKGuby0hcs}
}

@inproceedings{
marks2024the,
title={The Geometry of Truth: Emergent Linear Structure in Large Language Model Representations of True/False Datasets},
author={Samuel Marks and Max Tegmark},
booktitle={First Conference on Language Modeling},
year={2024},
url={https://openreview.net/forum?id=aajyHYjjsk}
}

@inproceedings{azaria-mitchell-2023-internal,
    title = "The Internal State of an {LLM} Knows When It`s Lying",
    author = "Azaria, Amos  and
      Mitchell, Tom",
    editor = "Bouamor, Houda  and
      Pino, Juan  and
      Bali, Kalika",
    booktitle = "Findings of the Association for Computational Linguistics: EMNLP 2023",
    month = dec,
    year = "2023",
    address = "Singapore",
    publisher = "Association for Computational Linguistics",
    url = "https://aclanthology.org/2023.findings-emnlp.68/",
    doi = "10.18653/v1/2023.findings-emnlp.68",
    pages = "967--976"
}

@inproceedings{
li2023inferencetime,
title={Inference-Time Intervention: Eliciting Truthful Answers from a Language Model},
author={Kenneth Li and Oam Patel and Fernanda Vi{\'e}gas and Hanspeter Pfister and Martin Wattenberg},
booktitle={Thirty-seventh Conference on Neural Information Processing Systems},
year={2023},
url={https://openreview.net/forum?id=aLLuYpn83y}
}

@inproceedings{
yuksekgonul2024attention,
title={Attention Satisfies: A Constraint-Satisfaction Lens on Factual Errors of Language Models},
author={Mert Yuksekgonul and Varun Chandrasekaran and Erik Jones and Suriya Gunasekar and Ranjita Naik and Hamid Palangi and Ece Kamar and Besmira Nushi},
booktitle={The Twelfth International Conference on Learning Representations},
year={2024},
url={https://openreview.net/forum?id=gfFVATffPd}
}

@inproceedings{
ho2025novo,
title={NoVo: Norm Voting off Hallucinations with Attention Heads in Large Language Models},
author={Zheng Yi Ho and Siyuan Liang and Sen Zhang and Yibing Zhan and Dacheng Tao},
booktitle={The Thirteenth International Conference on Learning Representations},
year={2025},
url={https://openreview.net/forum?id=yaOe2xBcLC}
}

@inproceedings{
du2024haloscope,
title={HaloScope: Harnessing Unlabeled {LLM} Generations for Hallucination Detection},
author={Xuefeng Du and Chaowei Xiao and Yixuan Li},
booktitle={The Thirty-eighth Annual Conference on Neural Information Processing Systems},
year={2024},
url={https://openreview.net/forum?id=nfK0ZXFFSn}
}
